\documentclass{article} 
\usepackage{iclr2018_conference,times}
\usepackage{hyperref}
\usepackage{url}

\usepackage{amsmath}
\usepackage{amssymb}
\usepackage{amsthm}
\usepackage{caption}
\usepackage{subcaption}
\usepackage{graphicx}

\usepackage{tabularx}
\usepackage{booktabs} 
\newcommand\T{\rule{0pt}{2.6ex}}       

\newtheorem{theorem}{Theorem}[section]

\newtheorem{lemma}[theorem]{Lemma}
\theoremstyle{definition}
\newtheorem{definition}{Definition}[section]

\title{PAC-Bayesian Margin Bounds \\ for Convolutional Neural Networks}

\author{Konstantinos Pitas \& Pierre Vandergheynst \thanks{.} \\
LTS2\\
EPFL\\
Lausanne, Switzerland \\
\texttt{\{konstantinos.pitas,pierre.vandergheynst\}@epfl.ch} \\
\And
Mike Davies \\
IDCOM \\
University of Edinburgh \\
Edinburgh, UK \\
\texttt{\{mike.davies\}@ed.ac.uk} \\
}

\iclrfinalcopy 

\begin{document}

\maketitle

\begin{abstract}
Deep neural networks generalize well despite having more parameters than the number of training samples. This runs contrary to traditional learning theory intuition, and generalization error bounds obtained using traditional techniques for these classification architectures are vacuous. There have been a number of recent works using Margin and PAC-Bayes analyses trying to address this problem. We start from a recent bound based on PAC-Bayes theory for fully connected networks and extend it to the convolutional setting. Our bound is orders of magnitude better than the previous estimate, and is a step towards analysing the generalization behaviour of more realistic deep convolutional architectures. 
\end{abstract}

\section{Introduction}
Recent work by \citet{zhang2016understanding} showed experimentally that standard deep convolutional architectures can easily fit a random labelling over unstructured random noise. This challenged conventional wisdom that the good generalization of DNNs results from impicit regularization though the use of SGD or explicit regularization using dropout and batch normalization, and stirred a considerable amount of research towards rigorously understanding generalization for these now ubiquitous classification architectures.

A key idea dating back to \citet{hochreiter1997flat} has been that neural networks that generalize are flat minima in the optimisation landscape. This has been observed empirically in \citet{keskar2016large}, however in \citet{dinh2017sharp} the authors propose that the notion of flatness needs to be carefully defined. One usually proceeds by assuming that a flat minimum can be described with low precision while a sharp minimum requires high precision. This means that we can describe the statistical model corresponding to the flat minimum with few bits. Then one can use a minimum description length (MDL) argument to show generalization \citet{rissanen1983universal}. Alternatively one can quantitatively measure the "flatness" of a minimum by injecting noise to the network parameters and measuring the stability of the network output. The more a trained network output is stable to noise, the more "flat" the minimum to which it corresponds, and the better it's generalization \citet{shawe1997pac} \citet{mcallester1999some}. Using this measure of "flatness" \citet{neyshabur2017pac} proposed a GE bound for deep fully connected networks. 

The bound of \citet{neyshabur2017pac} depends linearly on the latent ambient dimensionality of the hidden layers. For convolutional architectures while the ambient dimensionality of the convolution operators is huge, the effective number of parameters is much smaller. Therefore with a careful analysis one can derive generalization bounds that depend on the intrinsic layer dimensionality leading to much tigher bounds. In a recent work \citet{arora2018stronger} the authors explore a similar idea. They first compress a neural network removing redundant parameters with little or no degradation in accuracy and then derive a generalization bound on the new compressed network. This results in much tighter bounds than previous works.

\textbf{Contributions of this paper.}
\begin{itemize}
\item We apply a sparsity based analysis on convolutional-like layers of DNNs. We define convolutional like layers as layers with a sparse banded structure similar to convolutions but without weight sharing. We show that the implicit sparsity of convolutional-like layers reduces significantly the capacity of the network resulting in a much tighter GE bound. 
\item We then extend our results to true convolutional layers, with weight sharing, finding again tighter GE bounds. Surprisingly the bound for convolutional-like and convolutional layers is of the same order of magnitude. 
\item For completeness we propose to sparsify fully connected layers in line with the work of \citet{arora2018stronger} in order to reduce the number of effective network parameters. We then derive improved GE bounds based on the new reduced parameters. 
\end{itemize}

\textbf{Other related works.}
A number of recent works have tried to analyze the generalization of deep neural networks. This include margin approaches such as \citet{sokolic2016robust} \citet{bartlett2017spectrally}. Other lines of inquiry have been investigating the data dependent stability of SGD \citet{kuzborskij2017data} as well as the implicit bias of SGD over separable data \citet{soudry2017implicit} \citet{neyshabur2017geometry}.

\section{Preliminaries}
We now expand on the PAC-Bayes framework. Specifically let $f_{\boldsymbol{w}}$ be any predictor (not necessarily a neural network) learned from the training data and parameterized by $\boldsymbol{w}$. We assume a prior distribution $P$ over the parameters which should be a \textbf{proper} Bayesian prior and cannot depend on the training data. We also assume a posterior $\mathcal{Q}$ over the predictors of the form $f_{\boldsymbol{w}+\boldsymbol{u}}$, where $\boldsymbol{u}$ is a random variable whose distribution can depend on the training data.  Then with probability at least $1-\delta$ we get:

\begin{equation}
\mathbb{E}_{\boldsymbol{u}} [ L_0(f_{\boldsymbol{w}+\boldsymbol{u}}) ] \leq \mathbb{E}_{\boldsymbol{u}}[\hat{L}_{0}(f_{\boldsymbol{w}+\boldsymbol{u}}) ] +\mathcal{O}(\sqrt{\frac{2KL(\boldsymbol{w}+\boldsymbol{u}||P) +\text{ln} \frac{2 m }{\delta} }{ m-1  }}) 
\end{equation}

Notice that the above gives a generalization result over a distribution of predictors. We now restate a usefull lemma from \citet{neyshabur2017pac} which can be used to give a generalization result for a single predictor instance.
\begin{lemma}
Let $f_{\boldsymbol{w}}(\boldsymbol{x}):\mathcal{X} \Rightarrow \mathbb{R}^k$ be any predictor (not necessarily a neural network) with parameters $\boldsymbol{w}$, and $P$ be any distribution on the parameters that is independent of the training data. Then with probability $\geq 1-\delta$ over the training set of size $m$, for any random pertubation $\boldsymbol{u}$ s.t. $\mathbb{P}_{\boldsymbol{u}}[\max_{\boldsymbol{x \in \mathcal{X}}} |f_{\boldsymbol{w}+\boldsymbol{u} }(\boldsymbol{x})-f_{\boldsymbol{w}}(\boldsymbol{x})|_2 \leq \frac{\gamma}{4} ] \geq \frac{1}{2}$, we have: 

\begin{equation}
L_0(f_{\boldsymbol{w}}) \leq \hat{L}_{\gamma}(f_{\boldsymbol{w}})+\mathcal{O}(\sqrt{\frac{KL(\boldsymbol{w}+\boldsymbol{u}||P) +\text{ln} \frac{6 m }{\delta} }{ m-1  }}) 
\end{equation}

where $\gamma,\delta > 0$ are constants.
\end{lemma}
Let's look at some intuition behind this bound. It links the empirical risk $\hat{L}_{\gamma}(f_{\boldsymbol{w}})$ of the predictor to the true risk $L_0(f_{\boldsymbol{w}})$, for a specific predictor and not a posterior distribution of predictors. We have also moved to using a margin $\gamma$ based loss, this is essential step in order to remove the posterior assumption. The pertubation $\boldsymbol{u}$ quantifies how the true risk would be affected by choosing a bad predictor. The condition $\mathbb{P}_{\boldsymbol{u}}[\max_{\boldsymbol{x \in \mathcal{X}}} |f_{\boldsymbol{w}+\boldsymbol{u} }(\boldsymbol{x})-f_{\boldsymbol{w}}(\boldsymbol{x})|_2 \leq \frac{\gamma}{4} ] \geq \frac{1}{2}$ can be interpreted as choosing a posterior with \textbf{small variance}, sufficiently concentrated around the current empirical estimate $\boldsymbol{w}$, so that we can remove the randomness assumption with high confidence.

How small should we choose the the variance of $\boldsymbol{u}$? The choice is complicated because the KL term in the bound is \textbf{inversely proportional} to the variance of the pertubation. Therefore we need to find the largest possible variance for which our stability condition holds.

\begin{figure}[h!]
\centering
\begin{subfigure}{.5\textwidth}
  \centering
  \includegraphics[scale = 0.3]{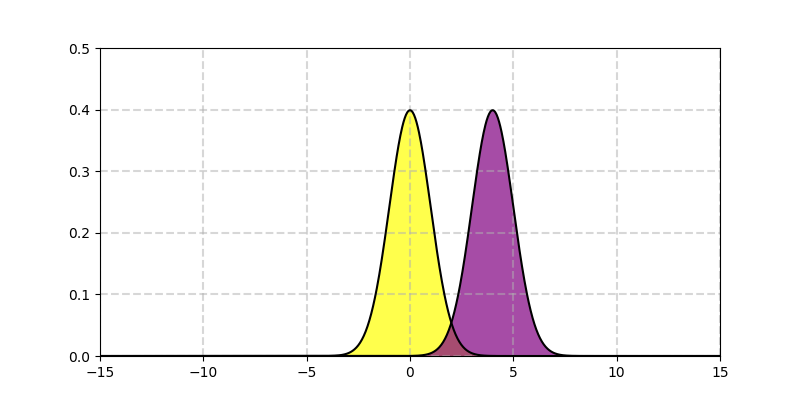}
  \caption{$\sigma = 1$ }
  \label{fig:sub1}
\end{subfigure}%
\begin{subfigure}{.5\textwidth}
  \centering
  \includegraphics[scale = 0.3]{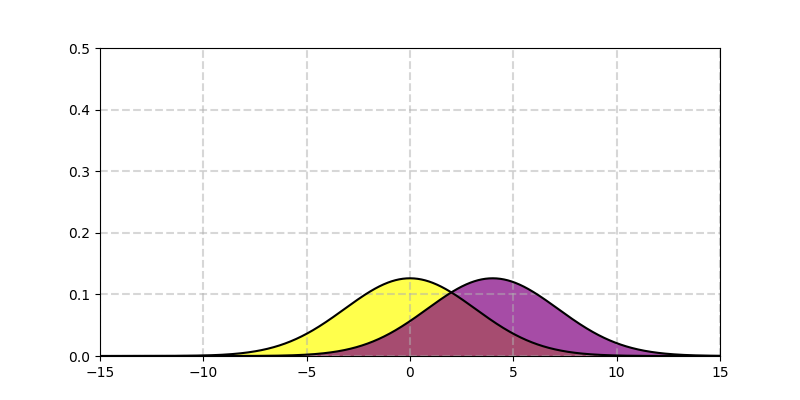}
  \caption{$\sigma = 10$}
  \label{fig:sub2}
\end{subfigure}
\caption{\textbf{Overlap between} $x_1 \sim \mathcal{N}(0,\sigma^2)$ \textbf{and} $x_2 \sim \mathcal{N}(4,\sigma^2)$: We see that as the variance increases the KL divergence between the two distributions decreases.}
\label{fig:test}
\end{figure}

The basis of our analysis is the following pertubation bound from \citet{neyshabur2017pac} on the output of a DNN:

\begin{lemma}
(Pertubation Bound). For any $B,d > 0$, let $f_w:\mathcal{X}_{B,n} \Rightarrow \mathbb{R}^k$ be a d-layer network with ReLU activations.. Then for any $\boldsymbol{w}$, and $\boldsymbol{x} \in \mathcal{X}_{B,n}$, and pertubation $\boldsymbol{u} = \text{vec}(\{\boldsymbol{U}_i \}^d_{i=1} )$ such that $||\boldsymbol{U}_i ||_2 \leq \frac{1}{d}||\boldsymbol{W}_i ||_2$, the change in the output of the network can be bounded as follows:

\begin{equation}
	|f_{\boldsymbol{w}+\boldsymbol{u} }(\boldsymbol{x})-f_{\boldsymbol{w}}(\boldsymbol{x})|_2 \leq e^2B \tilde{\beta}^{d-1} \sum_i ||\boldsymbol{U}_i||_2
\end{equation}
where $e$, $B$ and $\tilde{\beta}^{d-1}$ are considered as constants after an appropriate normalization of the layer weights. 
\end{lemma}

We note that correctly estimating the spectral norm of the pertubation at each layer is critical to obtaining a tight bound. Specifically if we exploit the structure of the pertubation we can \textbf{increase significantly} the variance of the added pertubation for which our stability condition holds.

We will also use the following definition of a network sparsification which will prove usefull later:

\begin{definition}
($\gamma,s$)-sparsification. Let $f$ be a classifier we say that $g_{s}$ is a ($\gamma,s$)-sparsification of $f$ if for any $x$ in the training set $S$ we have for all labels F$y$

\begin{equation}
|f(x)[y] -g_s(x)[y]|\leq \gamma
\end{equation}

and each fully connected layer in $g_{s}$ has at most $s$ non-zero entries along each row and column.
\end{definition}

We will assume for simplicity that we can always obtain a ($0,s$)-sparsification of $f$ see for example \citet{han2015learning}\citet{dong2017learning}\citet{pitas2018feta} and as such will therefore use $f$ and $g_{s}$ interchangeably.

\section{Layerwise Pertubations}
We need to find the maximum variance for which $\mathbb{P}_{\boldsymbol{u}}[\max_{\boldsymbol{x \in \mathcal{X}}} |f_{\boldsymbol{w}+\boldsymbol{u} }(\boldsymbol{x})-f_{\boldsymbol{w}}(\boldsymbol{x})|_2 \leq \frac{\gamma}{4} ] \geq \frac{1}{2}$. For this we present the following lemmas which bound the spectral norm of the noise at each layer. Specifically we assume a variance level $\sigma^2$ for the noise applied to each DNN parameter, based on the sparsity structure of a given layer we obtain noise matrices $\boldsymbol{U}$ with different structure and corresponding concentration inequalitites of the spectral norm $||\boldsymbol{U}||_2$. In the following we omit log parameters for clarity. 

\subsection{Fully Connected Layers}
We first start with fully connected layers that have been sparsified to a sparsity level $s$. After some calculations we get the following: 
\begin{theorem}
Let $\boldsymbol{U} \in \mathbb{R}^{d_2 \times d_1}$ be the pertubation matrix of a fully connected layer with with row and column sparsity equal to $s$. Then for $\boldsymbol{u} \sim \mathcal{N}(0,\sigma^2 \boldsymbol{I})$ the spectral norm of $\boldsymbol{U}$ behaves like:


\begin{equation}
\mathbb{P}(||\boldsymbol{U}||_2  \geq \sigma \{ 2\sqrt{s} + t\}) \leq e^{-\frac{t^2}{2}}.
\end{equation}

\end{theorem}

Note that in the above theorem we for fully connected layers without sparsity we can set $s = \max(d_1,d_2)$.

\begin{figure*}[t!]
\centering
\begin{subfigure}{.5\textwidth}
  \centering
  \includegraphics[scale=0.5]{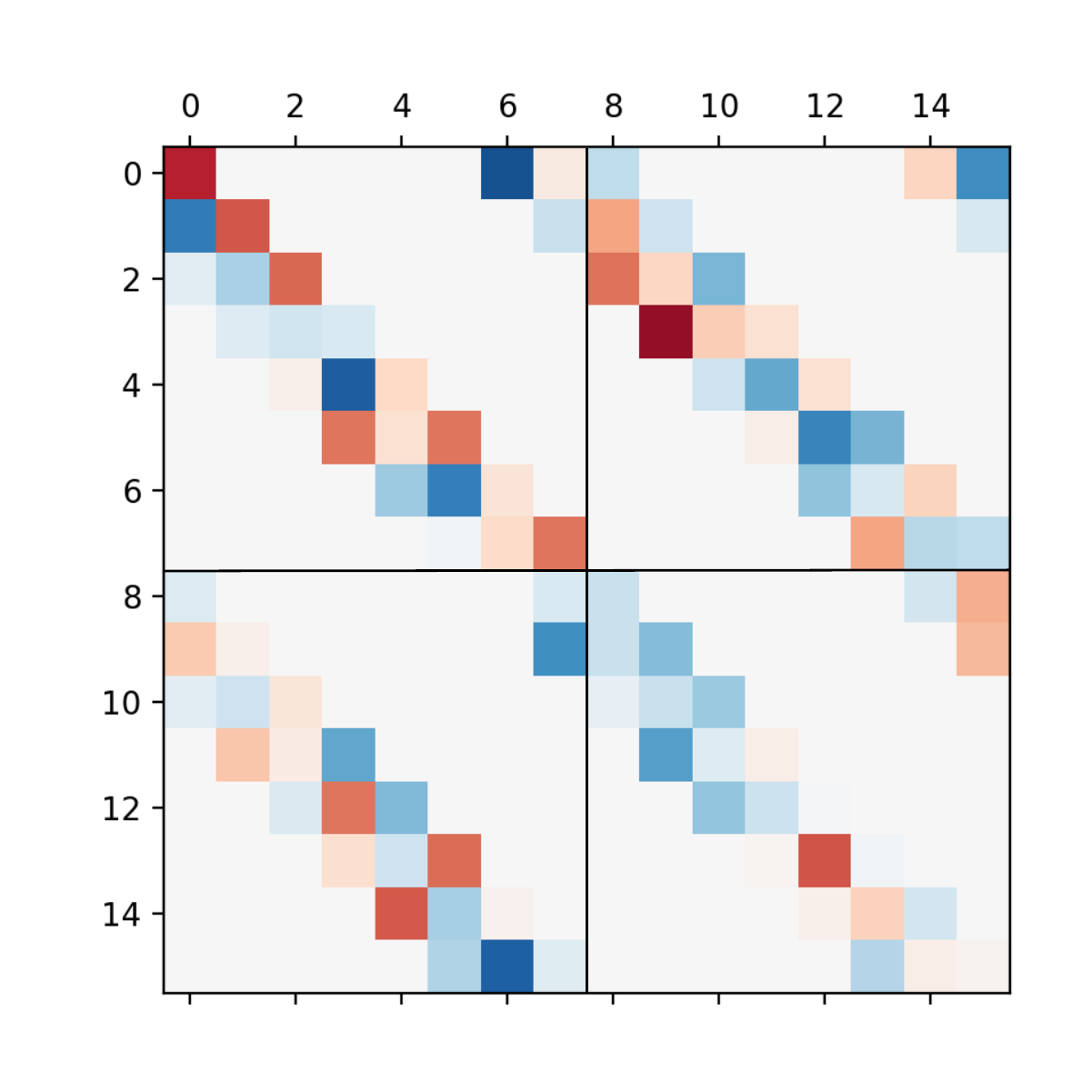}
  \caption{}
\end{subfigure}%
\begin{subfigure}{.5\textwidth}
  \centering
  \includegraphics[scale=0.5]{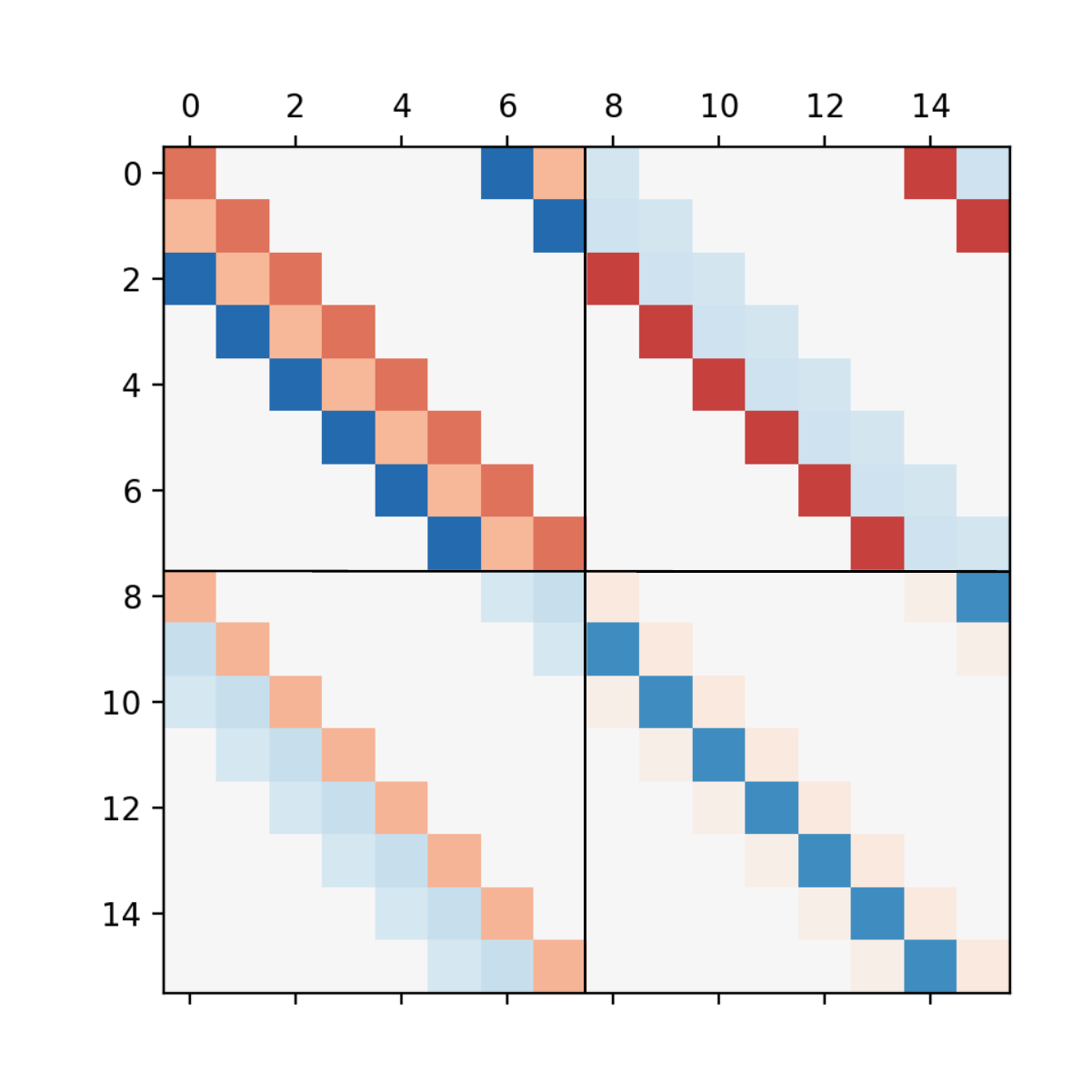}
  \caption{}
\end{subfigure}
\caption{\textbf{Structured Layers}: We separate two cases in our analysis a) Convolutional-Like layers where the matrix has a sparse banded structure but there is no weight sharing. b) Convolutional layers where the matrix represents convolutions and the weights are shared.}
\end{figure*}

\subsection{Convolutional-Like Layers}
We now analyze convolutional-like layers. These are layers that have a sparse banded structure similar to convolutions, with the simplifying assumption that the weights of the translated filters are not shared. This results in a matrix which for $1$d convolutions is plotted in Figure 2.a. While this type of layer is purely theoretical and does not directly correspond to practically applied layers, it is nevertheless usefull for isolating the effect of sparsity on the generalization error. After some calculations we get the following:
\begin{theorem}
Let $\boldsymbol{U} \in \mathbb{R}^{d_2 \times d_1}$ be the pertubation matrix of a 2$d$ convolutional-like layer with $a$ input channels, $b$ output channels, and convolutional filters $f \in \mathbb{R}^{q \times q}$. For $\boldsymbol{u} \sim \mathcal{N}(0,\sigma^2 \boldsymbol{I})$ and  the spectral norm of $\boldsymbol{U}$ behaves like:


\begin{equation}
\mathbb{P}(||\boldsymbol{U}||_2 \geq \sigma \{ q [\sqrt{a} + \sqrt{b}] + t \}   ) \leq e^{-\frac{t^2}{2}}.
\end{equation}
\end{theorem}

We see that the spectral norm of the noise is up to log parameters independent of the dimensions of the latent feature maps, and the ambient layer dimensionality. The spectral norm is a function of the root of the filter support $q$, the number of input channels $a$ and the number of output channels $b$.  

\subsection{Convolutional Layers}
We extend our analysis to true 2$d$ convolutional layers. After some calculations we get the following:
\begin{theorem}
Let $\boldsymbol{U} \in \mathbb{R}^{d_2 \times d_1}$ be the pertubation matrix of a 2$d$ convolutional layer with $a$ input channels, $b$ output channels, convolutional filters $f \in \mathbb{R}^{q \times q}$ and feature maps $F \in \mathbb{R}^{N \times N}$. For $\boldsymbol{u} \sim \mathcal{N}(0,\sigma^2 \boldsymbol{I})$ and  the spectral norm of $\boldsymbol{U}$ behaves like:


\begin{equation}
\mathbb{P}(||\boldsymbol{U}||_2 \geq \sigma  \{q [\sqrt{a} + \sqrt{b}] + t \}   ) \leq 2N^2e^{-\frac{t^2}{2q^2}}.
\end{equation}
\end{theorem}

We see that again up to log parameters, the spectral norm of the noise is independent of the dimensions of the latent feature maps. The spectral norm is a function of the root of the filter support $q$, the number of input channels $a$ and the number of output channels $b$.  The factor $2N^2$ on the concentration probability implies a less tight concentration than the convolutional-like layer. This is to be expected as the layer has fewer parameters due to weight sharing. We see also that up to log parameters the expected value of the spectral norms is the same for convolutional-like and convolutional layers. This is somewhat surprising as it implies similar generalization error bounds with and without weight sharing, contrary to conventional wisdom about DNN design. 

\section{Generalization Bound}

We now proceed to find the maximum value of the variance parameter $\sigma^2$. For this we use the following lemma:

\begin{lemma}
(Pertubation Bound). For any $B,d > 0$, let $f_w:\mathcal{X}_{B,n} \Rightarrow \mathbb{R}^k$ be a d-layer network with ReLU activations and we denote $C$ the set of convolutional layers and $D$ the set of dense layers. Then for any $\boldsymbol{w}$, and $\boldsymbol{x} \in \mathcal{X}_{B,n}$, and a pertubation for $\boldsymbol{u} \sim \mathcal{N}(0,\sigma^2 \boldsymbol{I})$, for any $\gamma > 0$ with 

\begin{equation}
\sigma  = \frac{\gamma}{42 B \tilde{\beta}^{d-1} [\sum_{i \in C} q_i \{\sqrt{a_i}+\sqrt{b_i}+\sqrt{2\log(4N_i^2d)} \} + \sum_{i \in D} \{2\sqrt{s_i}+\sqrt{2\log(2d)} \}]}  
\end{equation}

we have:

\begin{equation}
	\mathbb{P}_{\boldsymbol{u}}[\max_{\boldsymbol{x \in \mathcal{X}}} |f_{\boldsymbol{w}+\boldsymbol{u} }(\boldsymbol{x})-f_{\boldsymbol{w}}(\boldsymbol{x})|_2 \leq \frac{\gamma}{4} ] \geq \frac{1}{2}
\end{equation}
where $e$, $B$ and $\tilde{\beta}^{d-1}$ are considered as constants after an appropriate normalization of the layer weights. 
\end{lemma}

While we have deferred all other proofs to the Appendix we will describe the proof to this lemma in detail as it is crucial for understanding the GE bound.

\begin{proof}
We denote $C$ the set of convolutional layers, $D$ the set of dense layers and assume $|C|+|D|=d$ where $d$ is the total number of layers. We then assume that the probability for each of the $|D|$ events (5) and $|C|$ events (7) is upper bounded by $\frac{1}{2d}$. We take a union bound over these events and after some calculations obtain that:

\begin{equation}
\mathbb{P}(\sum_{i} ||\boldsymbol{U_i}||_2 \leq \sigma[\sum_{i \in C} q_i \{\sqrt{a_i}+\sqrt{b_i}+\sqrt{2\log(4N_i^2d)} \} + \sum_{i \in D} \{2\sqrt{s_i}+\sqrt{2\log(2d)} \}]) \geq \frac{1}{2}
\end{equation}

We are then ready to apply our result directly to Lemma 2.2. We calculate that with probability $\geq \frac{1}{2}$:

\begin{equation}
\begin{split}
&|f_{\boldsymbol{w}+\boldsymbol{u} }(\boldsymbol{x})-f_{\boldsymbol{w}}(\boldsymbol{x})|_2 \leq e^2B \tilde{\beta}^{d-1} \sum_i ||\boldsymbol{U}_i||_2 \\
&\leq \sigma e^2B \tilde{\beta}^{d-1} [\sum_{i \in C} q_i \{\sqrt{a_i}+\sqrt{b_i}+\sqrt{2\log(4N_i^2d)} \} + \sum_{i \in D} \{2\sqrt{s_i}+\sqrt{2\log(2d)} \}] \\
\end{split}
\end{equation}

We have now found a bound on the pertubation at the final layer of the network as a function of $\sigma$ with probability $\geq \frac{1}{2}$. What remains is to find the specific value of $\sigma$ such that $|f_{\boldsymbol{w}+\boldsymbol{u} }(\boldsymbol{x})-f_{\boldsymbol{w}}(\boldsymbol{x})|_2 \leq \frac{\gamma}{4}$. We calculate:

\begin{equation}
\begin{split}
&|f_{\boldsymbol{w}+\boldsymbol{u} }(\boldsymbol{x})-f_{\boldsymbol{w}}(\boldsymbol{x})|_2 \leq \frac{\gamma}{4} \\
&\Rightarrow \sigma e^2B \tilde{\beta}^{d-1} [\sum_{i \in C} q_i \{\sqrt{a_i}+\sqrt{b_i}+\sqrt{2\log(4N_i^2d)} \} + \sum_{i \in D} \{2\sqrt{s_i}+\sqrt{2\log(2d)} \}] \leq \frac{\gamma}{4}\\
&\Rightarrow \sigma \leq \frac{\gamma}{42 B \tilde{\beta}^{d-1} [\sum_{i \in C} q_i \{\sqrt{a_i}+\sqrt{b_i}+\sqrt{2\log(4N_i^2d)} \} + \sum_{i \in D} \{2\sqrt{s_i}+\sqrt{2\log(2d)} \}]}\\
\end{split}
\end{equation}
\end{proof}

We are now ready to state our main result. It follows directly from calculating the KL term in Lemma 2.1 where $\boldsymbol{w}+\boldsymbol{u} \sim \mathcal{N}(\boldsymbol{w},\sigma^2\boldsymbol{I})$ and $P \sim \mathcal{N}(0,\sigma^2\boldsymbol{I})$. For the variance value $\sigma^2$ chosen in equation (8):
\begin{theorem}
(Generalization Bound). For any  $B ,d ,h >0$, let $f_w:\mathcal{X}_{B,n} \Rightarrow \mathbb{R}^k$ be a d-layer network with ReLU activations. Then for any $\gamma,\delta > 0$, with probability $\geq 1-\delta$ over the training set of size $m$ we have: 

\begin{equation}
L_0(f_{\boldsymbol{w}}) \leq \hat{L}_{\gamma}(f_{\boldsymbol{w}})+\mathcal{O}(\sqrt{\frac{B^2 C_1^2 \prod_{i=1}^{k} ||\boldsymbol{W}_i||_2^2 \sum_{i=1}^k \frac{||\boldsymbol{W}_i||_2^F}{||\boldsymbol{W}_i||_2^2} }{ \gamma^2 m  }}) 
\end{equation}

where $C_1 = [\sum_{i \in C} q_i \{\sqrt{a_i}+\sqrt{b_i}+\sqrt{2\log(4N_i^2d)} \} + \sum_{i \in D} \{2\sqrt{s_i}+\sqrt{2\log(2d)} \}] $.
\end{theorem}
We can now compare this result from the previous by \citet{neyshabur2017pac}. For the case where all layers are sparse with sparsity $s$ ignoring log factors our bound scales like $C_1 = \mathcal{O}(d^2 s)$. For the case where all the layers are convolutional with $a_i = b_i = a$  ignoring log factors our bound scales like $C_1 = \mathcal{O}(d^2 q^2 a)$. For the same cases \citet{neyshabur2017pac} scales like $C_1 = \mathcal{O}(d^2h)$. We see that for convolutional layers $q^2 a \ll h$ our bound is orders of magnitude tighter for convolutional layers. Similarly for sparse fully connected layers $s \ll h$.

\section{Experiments}

\subsection{Concentration Bounds}
In this section we present experiments that validate the proposed concentration bounds for convolutional-like and convolutional layers. We assume $1$d convolutions, $a$ input channels, $b$ output channels and calculate theoretically and experimentally the spectral norm $||\boldsymbol{U}||_2$ of the random pertubation matrix $\boldsymbol{U}$. We increase the number of input and output channels assuming that $\tilde{a} = a = b$. To find empirical estimates we average the results over $N=100$ iterations for each choice of $\tilde{a}$. We plot the results in Figure 3. We see that the results for the expected value deviate by some log factors while the bounds correctly capture the growth rate of the expected spectral norm $||\boldsymbol{U}||_2$ as the parameters $a,b$ increase. Furthermore the empirical estimates validate the prediction that the norm $||\boldsymbol{U}||_2$ will be less concentrated around the mean for the true convolutional layer. 

\begin{figure}[h!]
\centering
\includegraphics[scale = 0.7]{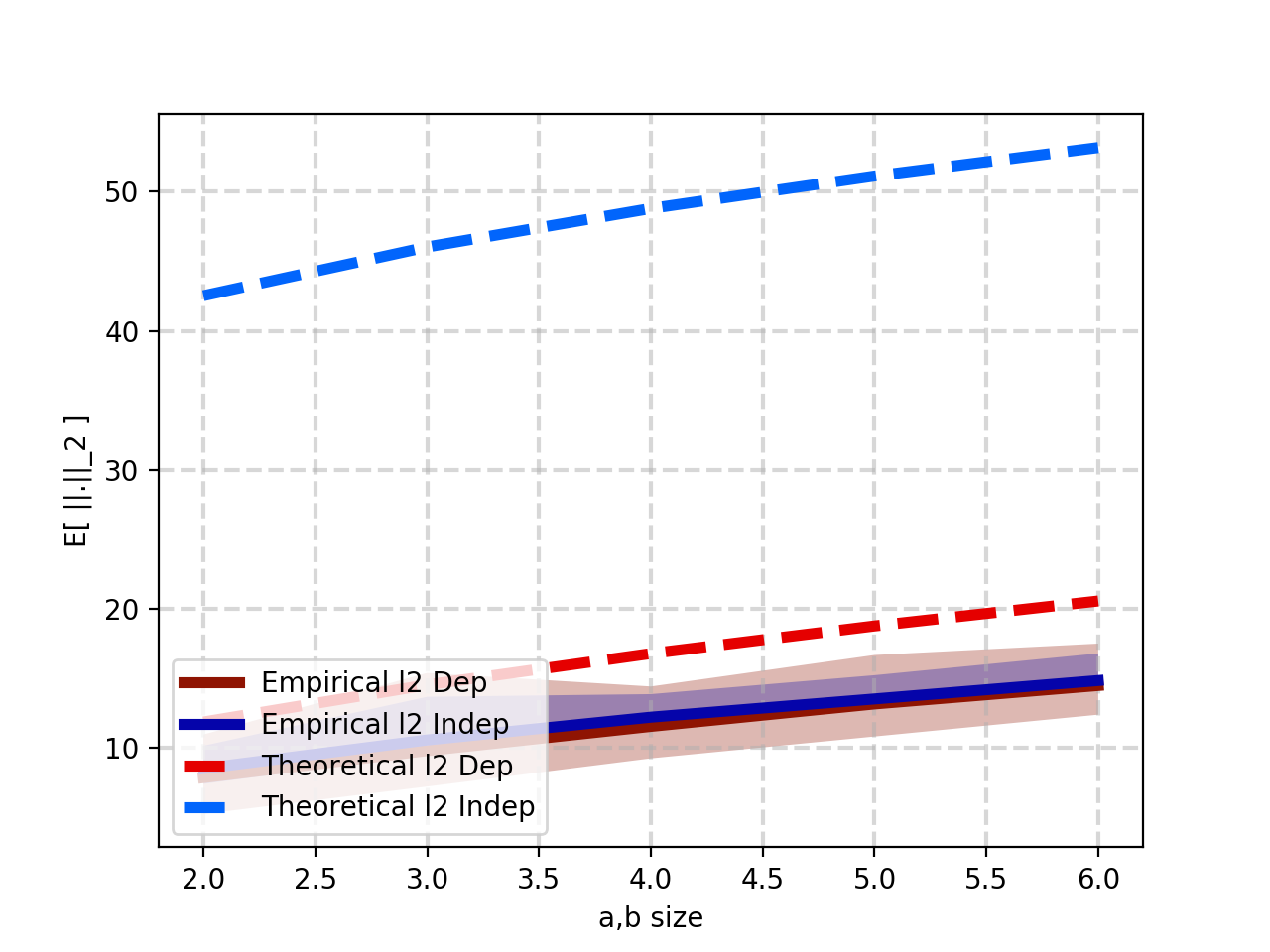}
\caption{\textbf{Empirical vs Theoretical Spetral Norms}: We plot the empirical (solid lines) and theoretical (dashed lines) estimates for $\mathbb{E}[ ||\boldsymbol{U}||_2 ]$. We also shade the area between the maximum and minimum empirical values. We see that bounds are tight up to log factors and correctly capture the norm growth as the number of channels increases. }
\end{figure} 

\subsection{Generalization Error Bounds}
For each layer we define $C_i$ the expected value of $||\boldsymbol{U}_i||_2$ where $C_i \in \{ \sqrt{d_{i1}}+\sqrt{d_{i2}}, 2\sqrt{s_i} , q_i(\sqrt{a_i}+\sqrt{b_i}) \}$ depending on different estimations of $\mathbb{E}[ ||\boldsymbol{U}_i||_2 ]$. We then plot $C_i^2$ on \textit{log} scale for different layers of some common DNN architectures. The intuition behind this plot is that if we consider a network where all layers are the same i.e. $i_{\star}$, the bound of (13) scales like $C_1^2 = d^2C_{i_{\star}}^2$. We experiment on LeNet-5 for the MNIST dataset, and on AlexNet and VGG-16 for the Imagenet dataset. We see that for convolutional layers the original approach of \citet{neyshabur2017pac} gives too pessimistic estimates, orders of magnitude higher than our approach. 

We see also that for large $a$ and $b$ the convolutional-like estimate is approximately the same as the convolutional estimate. 

We take also a closer look at the sample complexity estimate for the above architecture. We assume that $||\boldsymbol{W}_i||_2 \approx 1 , \; \forall i \in \{1,...,K\}$. We plot the results in Table 1. We obtain results orders of magnitude tigher than the previous PAC-Bayesian approach.

\begin{table}[h!]
\caption{Generalization error bounds for common feedforward architectures.} \label{tab:title2} 
\label{sample-table}
\vskip 0.15in
\begin{center}
\begin{small}
\begin{sc}
\begin{tabular}{ lcccc  }
  \toprule
   & LeNet-5 & AlexNet & VGG-16 \\ 
  \midrule
  \citet{neyshabur2017pac} & $\mathcal{O}(\sqrt{\frac{10^5 }{ m }})$ & $\mathcal{O}(\sqrt{\frac{10^{7} }{ m }})$ & $\mathcal{O}(\sqrt{\frac{10^{8} }{ m }})$ \T\\		
  Ours  & $\mathcal{O}(\sqrt{\frac{10^4 }{ m }})$ & $\mathcal{O}(\sqrt{\frac{10^5 }{ m }})$ & $\mathcal{O}(\sqrt{\frac{10^5 }{ m }})$  \\
  \bottomrule
\end{tabular}
\end{sc}
\end{small}
\end{center}
\vskip -0.1in
\end{table}

An interesting observation its that a original approach estimates the sample complexity of VGG-16 to be 1 order of magnitude larger than AlexNet even though both are trained on the same dataset (Imagenet) and do not overfit. By contrast our bound estimates approaximately the same sample complexity for both architectures, even though they differ significantly is ambient architecture dimensions. We also observe that our bound results in improved estimates when the spatial support of the filters is small relative to the dimensions of the feature maps. We observe that for the LeNet-5 architecture where the size of the support of the filters is big relative to the ambient dimension the benefit from our approach is small as the convolutional layers can be adequately modeled as dense matrices. 

It must be noted here that the assumption $||\boldsymbol{W}_i||_2 \approx 1 , \; \forall i \in \{1,...,K\}$ is quite strong and the bound is still worse than naive parameter counting when applied to real trained networks.

\begin{figure}[h!]
\centering

\subcaptionbox{$C^2_i$ \textbf{constants for the LeNet-5 architecture.}}
{\includegraphics[scale = 0.5]{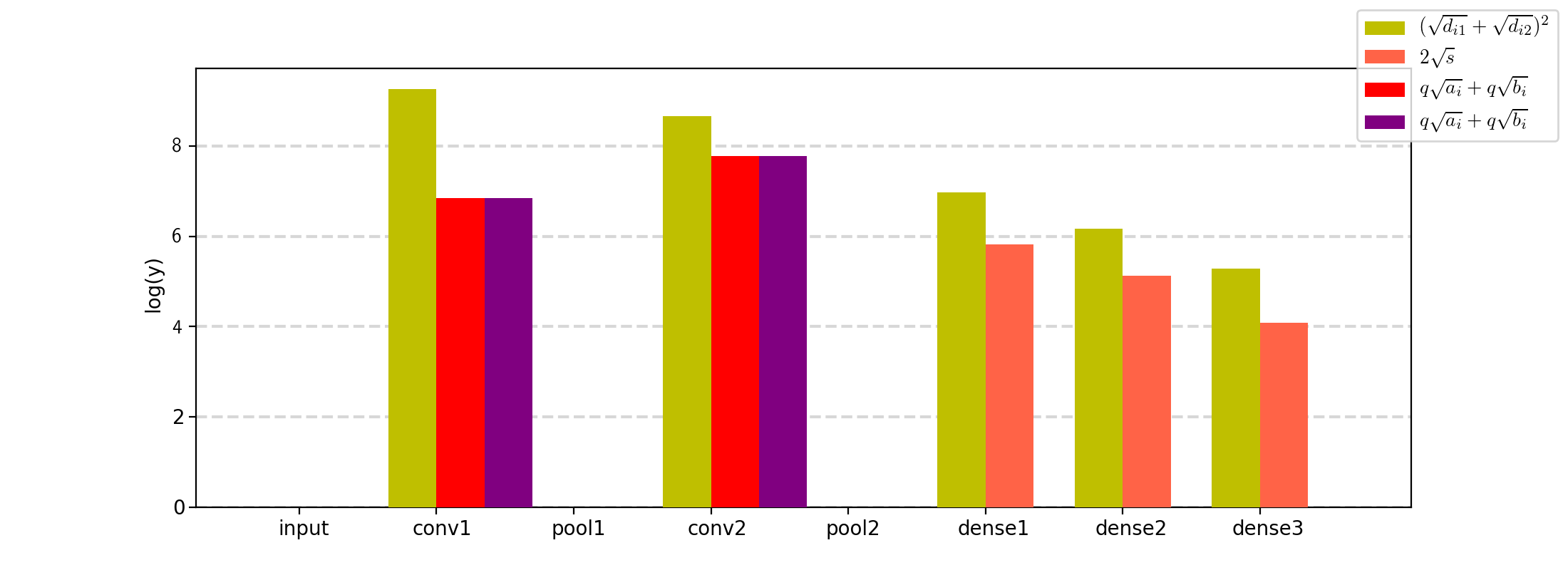}}

\subcaptionbox{$C^2_i$ \textbf{constants for the AlexNet architecture.}}
{\includegraphics[scale = 0.5]{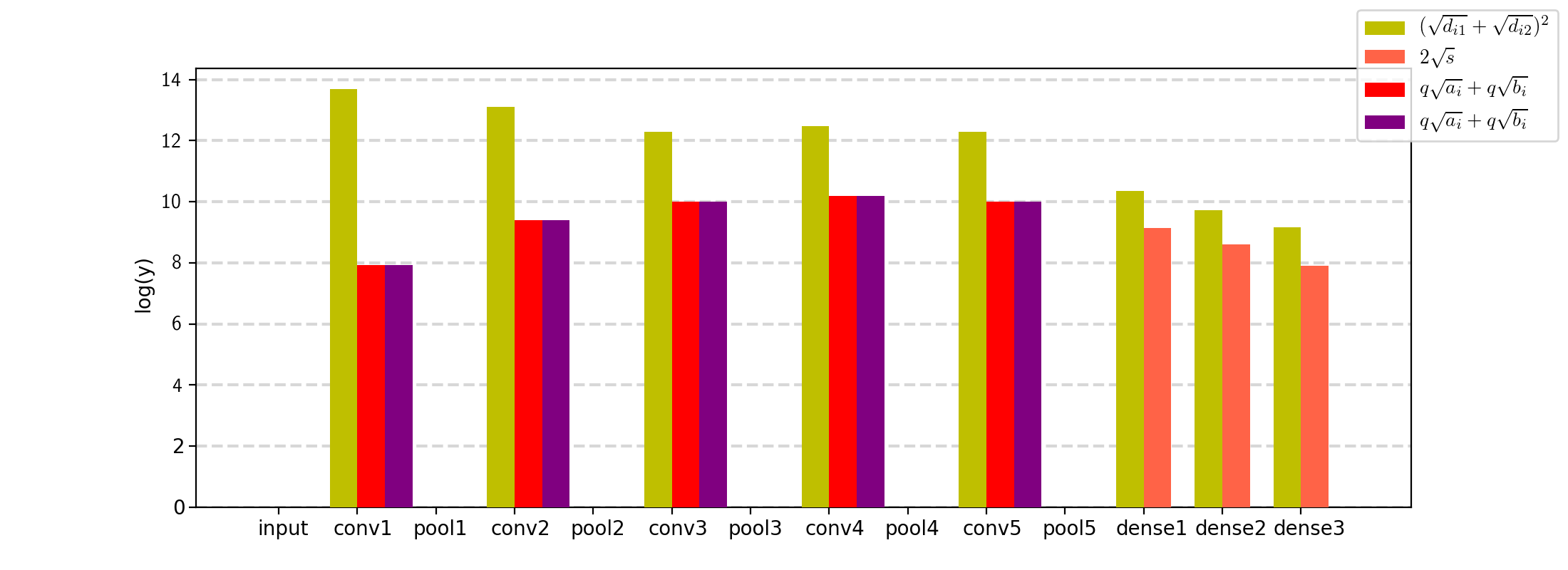}}

\subcaptionbox{$C^2_i$ \textbf{constants for the VGG-16 architecture.}}
{\includegraphics[scale = 0.5]{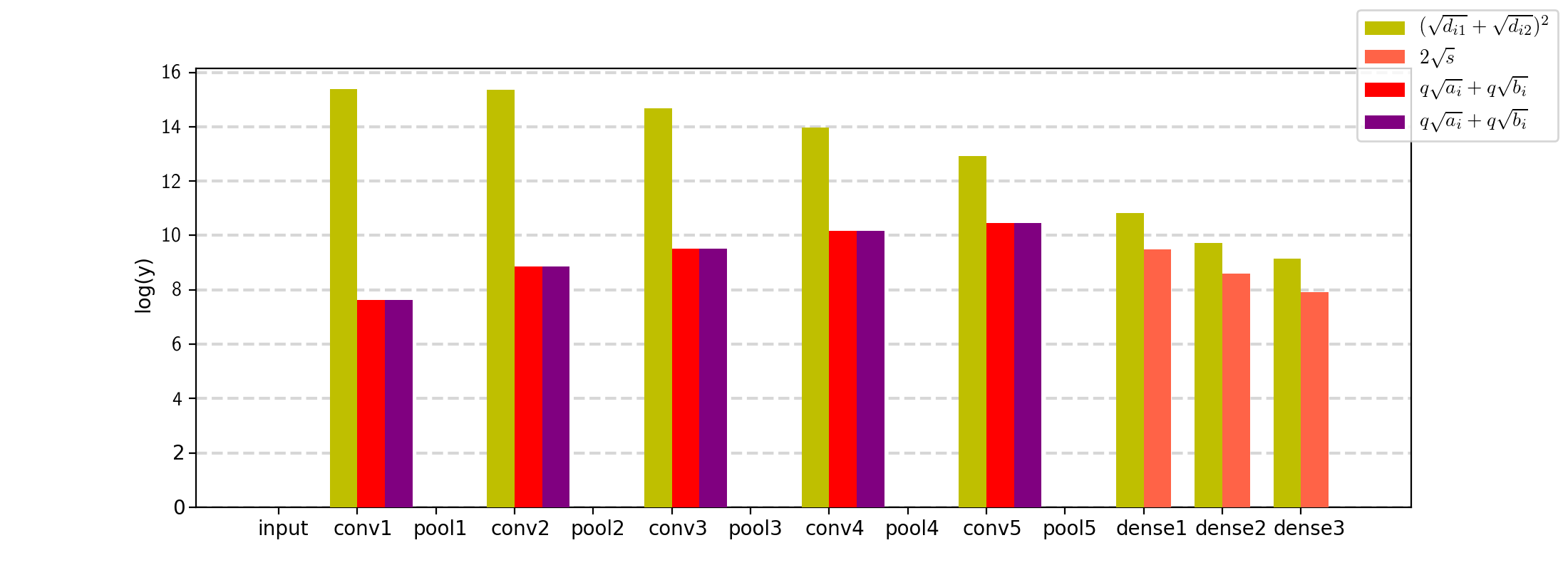}}

\caption{$C^2_i$ \textbf{constants for common feedforward architectures}: With yellow we plot the the estimation based on the ambient dimensionality. With magenta we plot the estimation based on sparsity, all layers have been compressed to $90\%$ sparsity. With red we plot the convolutional-like estimation. With purple we plot the convolutional estimation. We see that the layerwise $C^2_i$ constants are improved by orders of magnitude. Furthmore we notice that for convolutional layers the improvement is more significant for deeper architectures where the spatial support of the convolutional filters is much smaller than the size of the feature maps. For fully connected layers somewhat disappointingly the improvement is not very large. }
\end{figure}

\section{Discussion}
We have presented a new PAC-Bayes bound for deep convolutional neural networks. By decoupling the analysis from the ambient dimension of convolutional layers we manage to obtain bounds orders of magnitude tighter than existing approaches. We present numerical experiments on common feedforward architectures that corroborate our theoretical analysis.

\bibliography{iclr2018_conference}
\bibliographystyle{iclr2018_conference}
\clearpage

\section{APPENDIX}
In the derivations below we will rely upon two the following usefull theorem for the concentration of the spectral norm of sparse random matrices \citet{bandeira2016sharp}: 

\begin{theorem}
Let $\boldsymbol{A}$ be a $d_2 \times d_1$ random rectangular matrix with $\boldsymbol{A}_{ij} = \xi_{ij} \psi_{ij}$ where $ \{ \xi_{ij}:1 \leq i \leq d_2 , 1\leq j \leq d_1 \} $ are independent $ \mathcal{N}(0,1)$ random variables and $\{ \psi_{ij}:1 \leq i \leq d_2 , 1\leq j \leq d_1 \}$ are scalars. Then:

\begin{equation}
\mathbb{P}(||\boldsymbol{A}||_2 \geq (1+\epsilon) \{ \sigma_1 + \sigma_2 + \frac{5}{\sqrt{\log (1+\epsilon)} }\sigma_* \sqrt{\log (d_2 \wedge d_1)}  + t  \} ) \leq e^{-t^2 / 2 \sigma_*^2 }
\end{equation}

for any $0 \leq \epsilon \leq 1/2$ and  $t \geq 0$ with:

\begin{equation}
\sigma_1:= \max_i \sqrt{ \sum_j \psi_{ij}^2 } \qquad \sigma_2:= \max_i \sqrt{ \sum_j \psi_{ij}^2 } \qquad \sigma_*:= \max_{ij}|\psi_{ij}|.
\end{equation}
\end{theorem}

\subsection{Proof of Theorem 3.1}
For $\boldsymbol{u} \sim \mathcal{N}(0, I)$ we get the result trivially from Therorem 7.1 by assuming: 
\begin{equation}
\psi_{ij} = 
\begin{cases} 
      0 & (i,j)\notin \text{supp}(f) \\
      1 & (i,j)\in \text{supp}(f) \\
\end{cases}
\end{equation}
. We can extend the result to $\sigma > 0$ by considering that $||\sigma \boldsymbol{U}_i ||_2 = \sigma||\boldsymbol{U}_i ||_2$.

\subsection{Proof of Theorem 3.2}

\begin{proof}
We will consider first the case $\boldsymbol{u} \sim \mathcal{N}(0, I)$. A convolutional layer is characterised by it's output channels. For each output channel each input channel is convolved with an independent filter, the output of the output channel is then the sum of the results of these convolutions. We consider convolutional-like layers, i.e. the layers are banded but the entries are independent and there is no weight sharing. We plot for the case of one dimensional signals the implied structure.

\begin{figure}[h!]
\centering
\includegraphics[scale = 0.6]{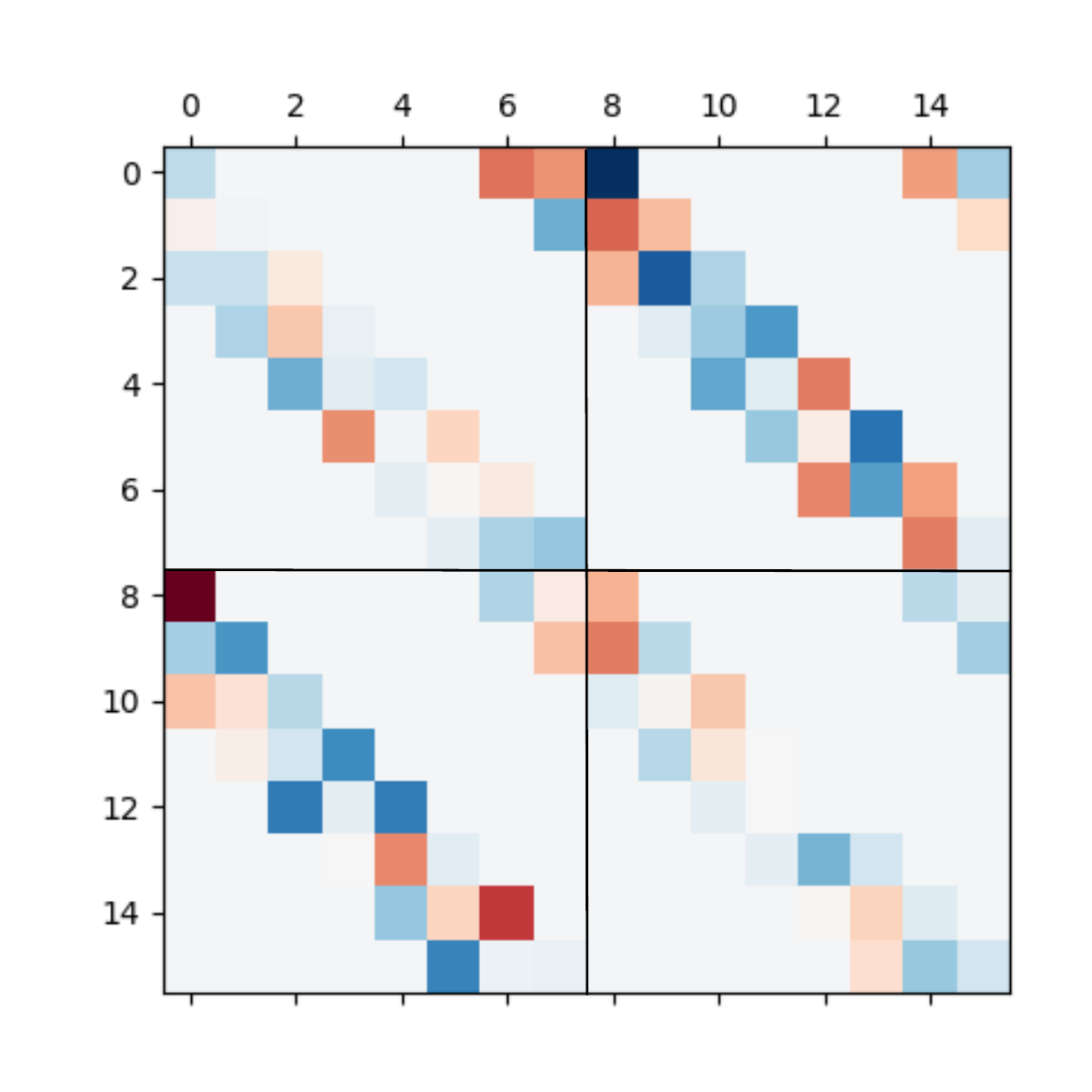}
\caption{Structure of a 1d convolutional-like layer with 2 input channels and 2 output channels.}
\end{figure}

We need to evaluate $\sigma_1:= \max_i \sqrt{ \sum_j \psi_{ij}^2 }$ , $\sigma_2:= \max_j \sqrt{ \sum_i \psi_{ij}^2 }$ and $\sigma_*:= \max_{ij}|\psi_{ij}|$ for a matrix this matrix. Where:

\begin{equation}
\psi_{ij} = 
\begin{cases} 
      0 & (i,j)\notin \text{supp}(f) \\
      1 & (i,j)\in \text{supp}(f) \\
\end{cases}
\end{equation}

Below we plot what these sums represent:

\begin{figure}[h!]
\centering
\begin{subfigure}{.5\textwidth}
  \centering
  \includegraphics[scale = 0.6]{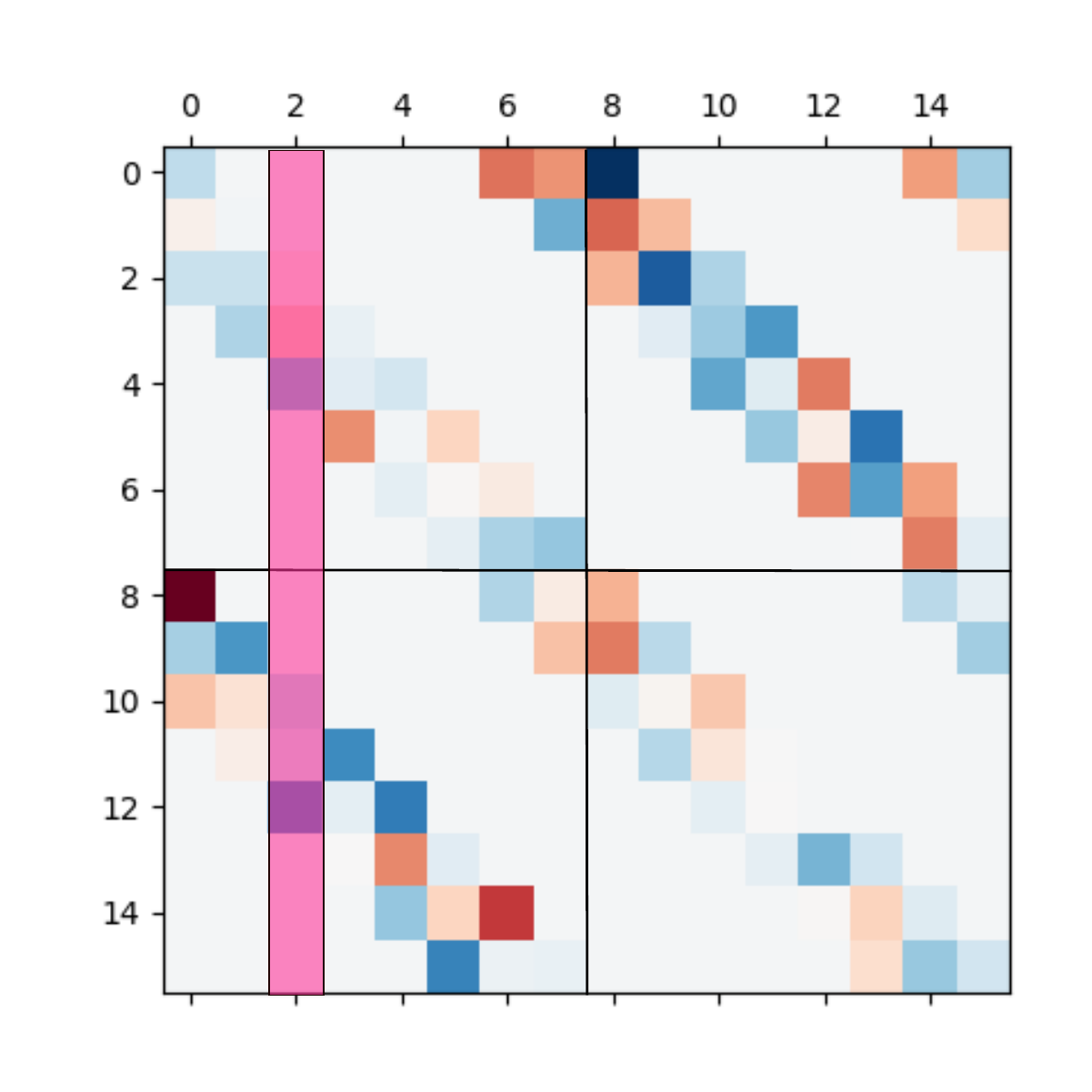}
  \caption{$\sigma_1:= \max_i \sqrt{ \sum_j \psi_{ij}^2 }$ }
  \label{fig:sub1}
\end{subfigure}%
\begin{subfigure}{.5\textwidth}
  \centering
  \includegraphics[scale = 0.6]{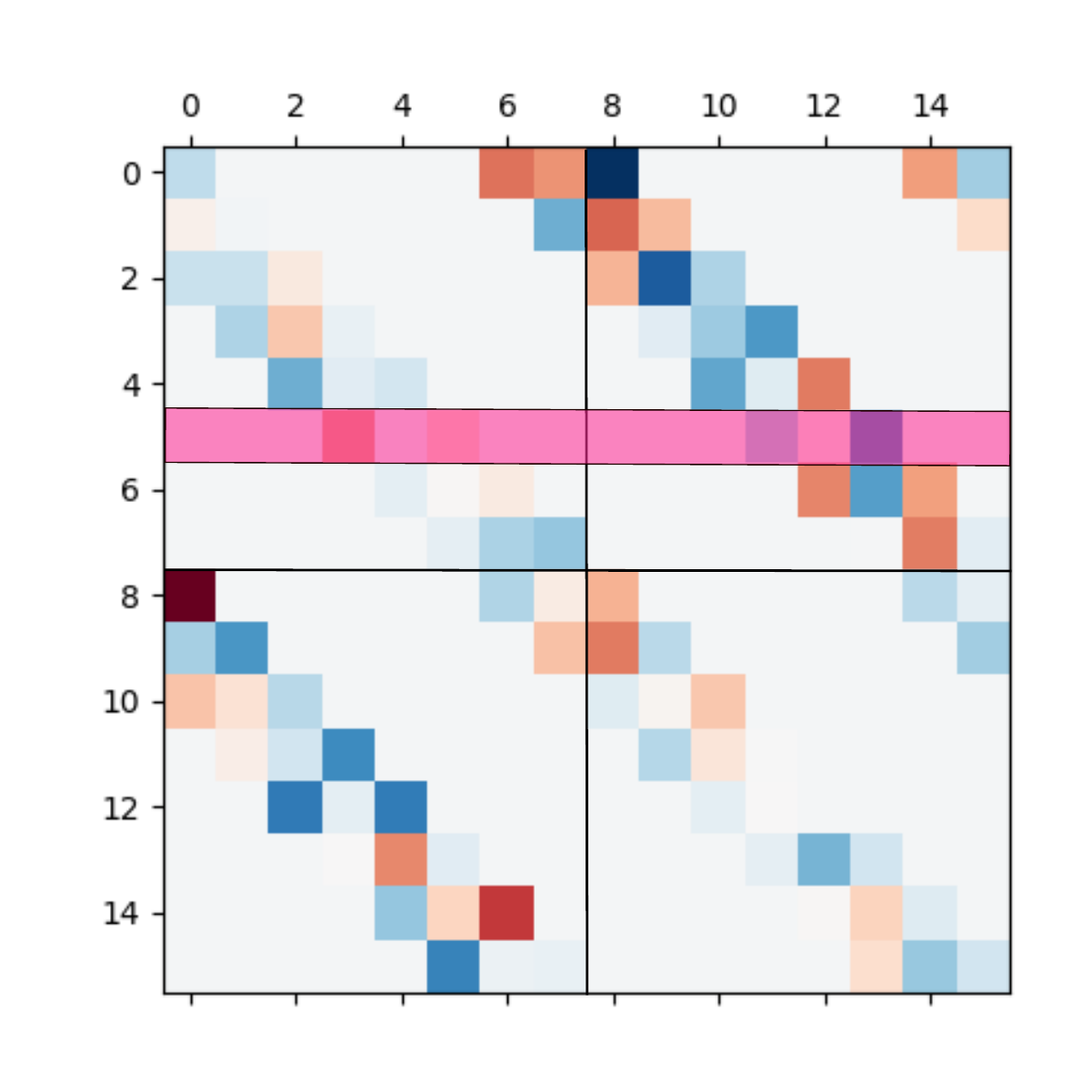}
  \caption{$\sigma_2:= \max_j \sqrt{ \sum_i \psi_{ij}^2 }$}
  \label{fig:sub2}
\end{subfigure}
\caption{$\sigma_1$ and $\sigma_2$}
\label{fig:test}
\end{figure}

For $\sigma_1$ we can find an upper bound, by considering that the sum for a given filter and a given pixel location represents the maximum number of overlaps for all 2d shifts. For the case of 2d this is $(q)^2$ equal to the support of the filters and we also need to consider all input channels. We then get 

\begin{equation}
\sigma_1 := \max_i \sqrt{ \sum_j \psi_{ij}^2 } \leq  \sqrt{\sum_{a_i}\sum_{q^2} \psi_{ij}^2 }  =  \sqrt{\sum_{a_i}\sum_{q^2} } =  \sqrt{a_iq^2} =  q\sqrt{a_i}  
\end{equation}

\begin{figure}[h!]
\includegraphics[scale = 0.3]{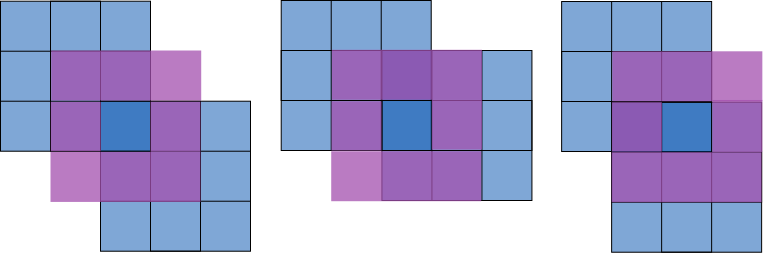}
\centering
\caption{Possible shifts with overlap: With blue we plot a 2d filter $f \in \mathbb{R}^{3\times3}$ and 3 filters $f \in \mathbb{R}^{3\times3}$ that overlap with it's bottom right pixel. In purple we plot the box denoting the boundaries of the set of all shifted filters that overlap with the bottom right pixel.} 

\end{figure}  

For $\sigma_2$ we need to consider that each column in the matrix represents a concatenation of convolutional filters $f \in \mathbb{R}^{q \times q} $. Then it is straight forward to derive that:

\begin{equation}
\sigma_1 := \max_i \sqrt{ \sum_j \psi_{ij}^2 } \leq  \sqrt{\sum_{b_i}\sum_{q^2} \psi_{ij}^2 }  =  \sqrt{\sum_{b_i}\sum_{q^2} } =  \sqrt{b_i q^2}  =  q\sqrt{b_i}  
\end{equation}

Furthermore it is trivial to show that $\sigma_* = 1$. The theorem results extends trivially to $\sigma > 0$ by considering that $||\sigma \boldsymbol{U}_i ||_2 = \sigma||\boldsymbol{U}_i ||_2$.
\end{proof}

\subsection{Proof of Theorem 3.3}

\begin{proof}

\begin{figure*}[t!]
\centering
\begin{subfigure}{.5\textwidth}
  \centering
  \includegraphics[scale=0.5]{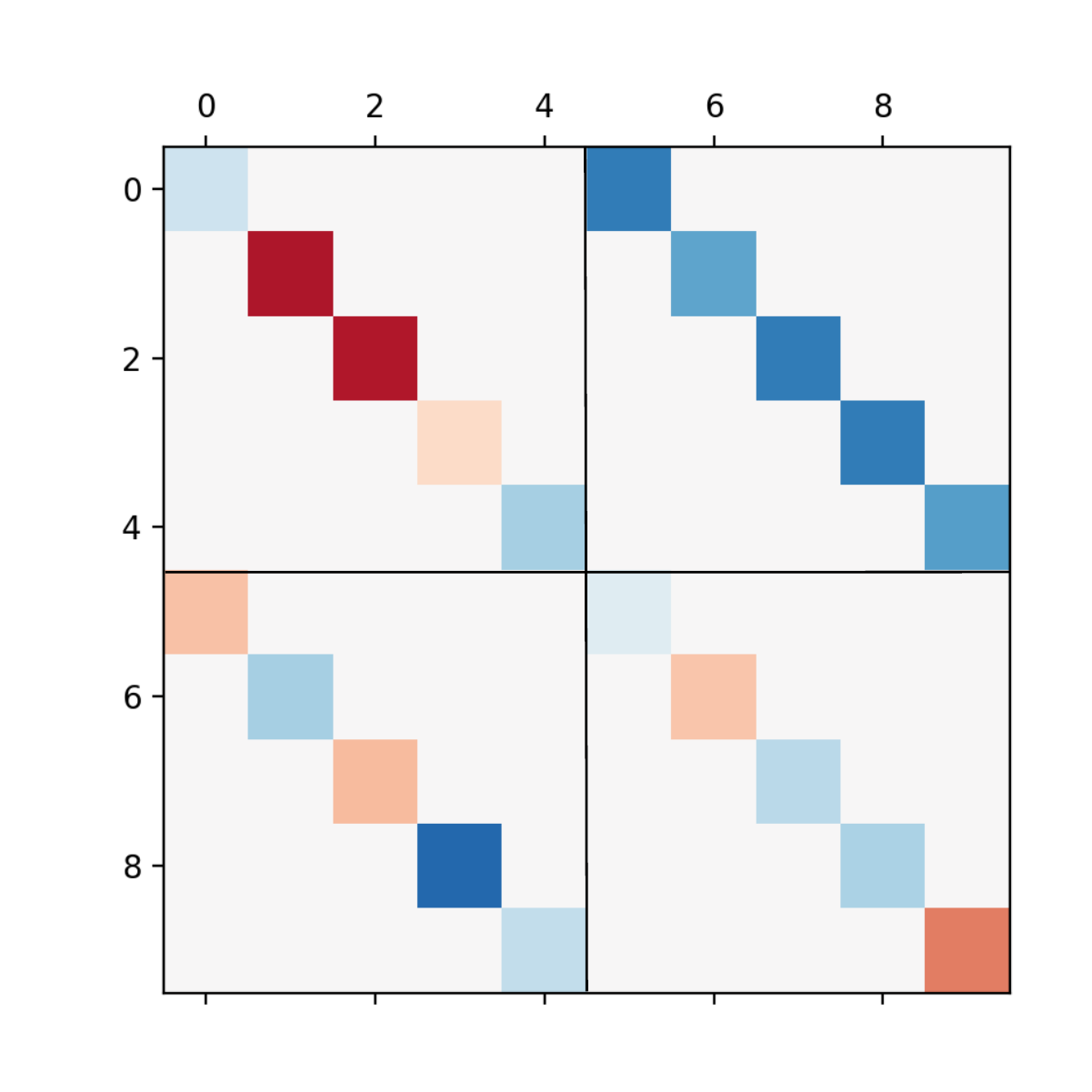}
  \caption{}
\end{subfigure}%
\begin{subfigure}{.5\textwidth}
  \centering
  \includegraphics[scale=0.5]{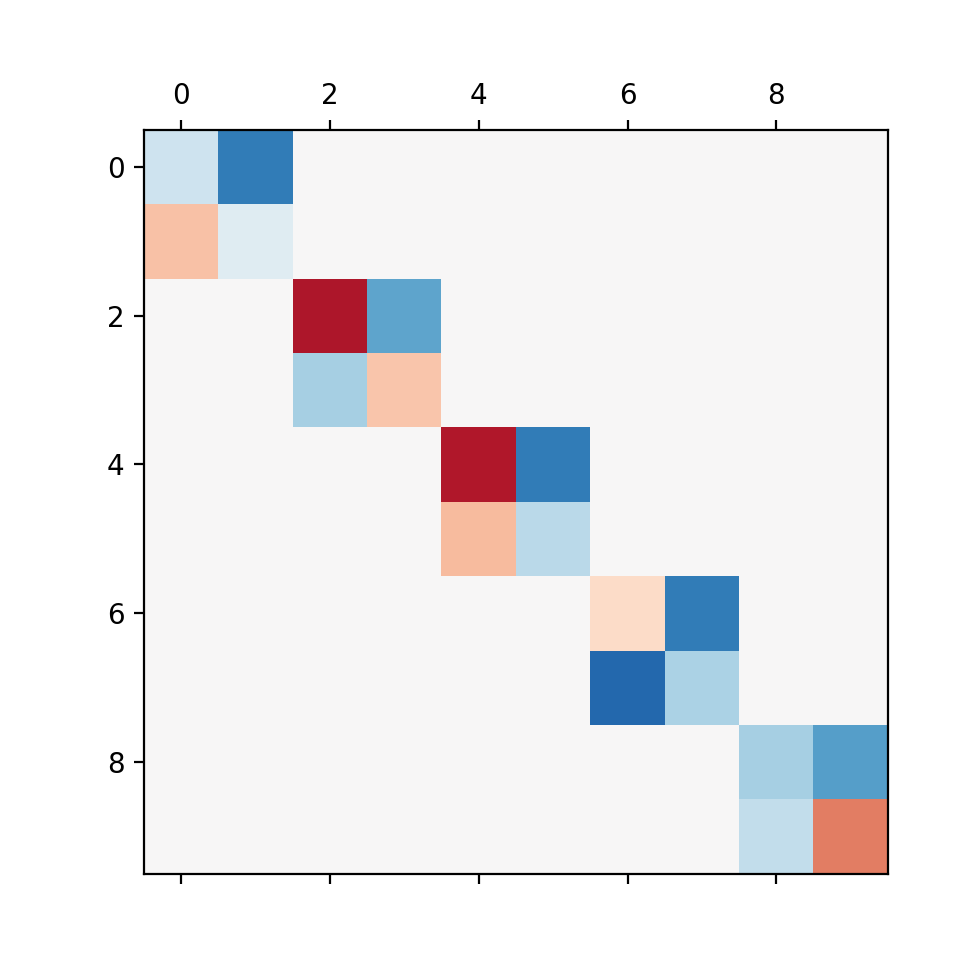}
  \caption{}
\end{subfigure}
\caption{\textbf{Concatenation of diagonal matrices}: We see that the concatenation of diagonal matrices can be always rearranged into a block diagonal matrix.}
\end{figure*}

We start by defining the structure of the 2d convolutional noise $\boldsymbol{U}$. Given $a$ input channels and $b$ output channels the noise matrix $\boldsymbol{U}$ is structured as:

\begin{equation}
\boldsymbol{U} = 
\begin{bmatrix}
    \boldsymbol{A}^{00}  & ... & \boldsymbol{A}^{0a}  \\
    \vdots  & \ddots & \vdots  \\
    \boldsymbol{A}^{b0}  & ... & \boldsymbol{A}^{ba}  \\
\end{bmatrix}
\end{equation}

We transform this matrix into the Fourier domain to obtain:

\begin{equation}
\begin{split}
||\boldsymbol{U}||_2 
& =
||
\begin{bmatrix}
    \boldsymbol{F}_{0}^{T}  & 0 & ... & 0  \\
    0  & \boldsymbol{F}_{1}^{T} & ... & 0  \\
    \vdots  & \vdots & \ddots & \vdots  \\
    0  & 0 & \hdots & \boldsymbol{F}_{b}^{T}  \\
\end{bmatrix}
\begin{bmatrix}
    \boldsymbol{\tilde{A} }^{00}  & ... & \boldsymbol{\tilde{A}}^{0a}  \\
    \vdots  & \ddots & \vdots  \\
    \boldsymbol{\tilde{A}}^{b0}  & ... & \boldsymbol{\tilde{A}}^{ba}  \\
\end{bmatrix}
\begin{bmatrix}
    \boldsymbol{F}_{0}  & 0 & ... & 0  \\
    0  & \boldsymbol{F}_{1} & ... & 0  \\
    \vdots  & \vdots & \ddots & \vdots  \\
    0  & 0 & \hdots & \boldsymbol{F}_{a}  \\
\end{bmatrix}||_2 \\
& =
||
\begin{bmatrix}
    \boldsymbol{F}_{0}^{T}  & 0 & ... & 0  \\
    0  & \boldsymbol{F}_{1}^{T} & ... & 0  \\
    \vdots  & \vdots & \ddots & \vdots  \\
    0  & 0 & \hdots & \boldsymbol{F}_{b}^{T}  \\
\end{bmatrix}
\begin{bmatrix}
    \boldsymbol{\tilde{B}}_{0}  & ... & 0  \\
    \vdots  & \ddots & \vdots  \\
    0  & ... & \boldsymbol{\tilde{B}}_{N^2}  \\
\end{bmatrix}
\begin{bmatrix}
    \boldsymbol{F}_{0}  & 0 & ... & 0  \\
    0  & \boldsymbol{F}_{1} & ... & 0  \\
    \vdots  & \vdots & \ddots & \vdots  \\
    0  & 0 & \hdots & \boldsymbol{F}_{a}  \\
\end{bmatrix}||_2 \\
& = 
||
\begin{bmatrix}
    \boldsymbol{\tilde{B}}_{0}  & ... & 0  \\
    \vdots  & \ddots & \vdots  \\
    0  & ... & \boldsymbol{\tilde{B}}_{N^2}  \\
\end{bmatrix} ||_2 
= \max_n\{ ||\boldsymbol{\tilde{B} }_n||_2 \} \leq \max_n\{ ||\text{Re}(\boldsymbol{\tilde{B}}_n)||_2 + ||\text{Im}(\boldsymbol{\tilde{B}}_n)||_2 \} \\
\end{split}
\end{equation}

 where:

\begin{equation}
\boldsymbol{\tilde{B}_{n}} = 
\begin{bmatrix}
    \lambda_n^{00}  & \hdots & \lambda_n^{0a}  \\
    \vdots  & \ddots & \hdots  \\
    \lambda_n^{b0}  & \hdots & \lambda_n^{ba}  \\
\end{bmatrix}
\end{equation}

and:

\begin{equation}
\lambda_n^{ij} = \lambda_{n_1,n_2}^{ij} = \sum_{k_1 = 0}^{K_1 - 1} \sum_{k_2 = 0}^{K_2 - 1} e^{-2 \pi i (\frac{k_1 n_1}{K_1}+\frac{k_2 n_2}{K_2})} \boldsymbol{\tilde{A}}^{ij}_{k_1,k_2}
\end{equation}

Then the matrices $\text{Re}(\boldsymbol{\tilde{B}}_n)$ and $\text{Im}(\boldsymbol{\tilde{B}}_n)$ have entries:

\begin{equation}
\begin{split}
& \text{Re}(\lambda_{n}^{ij}) \sim \mathcal{N}(0, \sum_{k_1 = 0}^{K_1 - 1} \sum_{k_2 = 0}^{K_2 - 1}\text{cos}^2(2 \pi i (\frac{k_1 n_1}{K_1}+\frac{k_2 n_2}{K_2})) \\
& \text{Im}(\lambda_{n}^{ij}) \sim \mathcal{N} (0, \sum_{k_1 = 0}^{K_1 - 1} \sum_{k_2 = 0}^{K_2 - 1}\text{sin}^2(2 \pi i (\frac{k_1 n_1}{K_1}+\frac{k_2 n_2}{K_2})) \\
\end{split}
\end{equation}.

In the first line we have used the fact that 2d convolution is diagonalized by the 2d Fourier transform into diagonal matrices $\boldsymbol{\tilde{A}_{ij}}$. In the second line we have used the fact that this concatenation of diagonal matrices can be always rearranged as a block diagonal matrix with blocks $\boldsymbol{\tilde{B}_{n}}$

We can now apply the following concentration inequality \citet{vershynin2010introduction}:
\begin{theorem}
Let $\boldsymbol{A}$ be an $N \times n$ matrix whose entries are independent standard normal random variables. Then for every $t \geq 0$:
\begin{equation}
\mathbb{P}(||\boldsymbol{A}||_2 \geq \sqrt{N} + \sqrt{n} + t) \leq e^{-t^2/2},
\end{equation}
\end{theorem}

on the matrices $\text{Re}(\boldsymbol{\tilde{B}}_n)$ and $\text{Im}(\boldsymbol{\tilde{B}}_n)$. We obtain the following concentration inequalities:

\begin{equation}
\begin{split}
& \mathbb{P}(|| \text{Re}(\boldsymbol{\tilde{B}_{n}}) ||_2 \geq \sigma_{re,n}(\sqrt{a}+\sqrt{b})+t ) \leq \exp(-t^2/2\sigma^2_{re,n}) \\
& \mathbb{P}(|| \text{Im}(\boldsymbol{\tilde{B}_{n}}) ||_2 \geq \sigma_{im,n}(\sqrt{a}+\sqrt{b})+t ) \leq \exp(-t^2/2\sigma^2_{im,n}) \\
\end{split}
\end{equation}

with:
\begin{equation}
\begin{split}
\sigma_{re,n} &= \sigma_{re,n_1,n_2} = \sqrt{ \sum_{k_1 = 0}^{K_1 - 1} \sum_{k_2 = 0}^{K_2 - 1}\text{cos}^2(2 \pi i (\frac{k_1 n_1}{K_1}+\frac{k_2 n_2}{K_2}) }\\
\sigma_{im,n} &= \sigma_{im,n_1,n_2} = \sqrt{ \sum_{k_1 = 0}^{K_1 - 1} \sum_{k_2 = 0}^{K_2 - 1}\text{sin}^2(2 \pi i (\frac{k_1 n_1}{K_1}+\frac{k_2 n_2}{K_2}) }.\\
\end{split}
\end{equation}

We then make the following calculations:

\begin{equation}
\begin{split}
\max_n [\sigma_{re,n}+\sigma_{im,n}] &= \max_n [\sqrt{ \sum_{k_1 = 0}^{K_1 - 1} \sum_{k_2 = 0}^{K_2 - 1}\text{cos}^2(2 \pi i (\frac{k_1 n_1}{K_1}+\frac{k_2 n_2}{K_2}) } \\
&+ \sqrt{ \sum_{k_1 = 0}^{K_1 - 1} \sum_{k_2 = 0}^{K_2 - 1}\text{sin}^2(2 \pi i (\frac{k_1 n_1}{K_1}+\frac{k_2 n_2}{K_2}) } ]\\
& \leq \sqrt{ \sum_{k_1 = 0}^{K_1 - 1} \sum_{k_2 = 0}^{K_2 - 1}\frac{1}{2}} + \sqrt{ \sum_{k_1 = 0}^{K_1 - 1} \sum_{k_2 = 0}^{K_2 - 1}\frac{1}{2} } = \frac{2}{\sqrt{2}}q \approx 1.4q
\end{split}
\end{equation}
since:

\begin{equation}
\begin{split}
& \frac{\partial}{\partial\theta_i}(\sqrt{\sum_i\sin^2(\theta_i)}+\sqrt{\sum_i\cos^2(\theta_i)}) = \frac{1}{2}\frac{2\cos(\theta_i)\sin(\theta_i)}{|\sin(\theta_i)|} - \frac{1}{2}\frac{2\cos(\theta_i)\sin(\theta_i)}{|\cos(\theta_i)|} \\
& \frac{\sin(\theta_i)\cos(\theta_i)}{|\sin(\theta_i)||\cos(\theta_i)|}(|\cos(\theta_i)|-|\sin(\theta_i)|) = 0 \\
& \Rightarrow \cos(\theta_i) = \sin(\theta_i) = \pm \frac{1}{\sqrt{2}}.
\end{split}
\end{equation}

We now notice that there are $N^2$ matrices $\text{Re}(\boldsymbol{\tilde{B}_{n}})$ and $N^2$ matrices $\text{Im}(\boldsymbol{\tilde{B}_{n}})$. We assume that the probability of each of the $N^2$ events in (21) is upper bounded by $\frac{T}{2N^2}$ for $T>0$ and take a union bound.

We then get:

\begin{equation}
\begin{split}
& \mathbb{P}(\bigcup_{n} \{ || \text{Re}(\boldsymbol{\tilde{B}_{n}}) ||_2 + || \text{Im}(\boldsymbol{\tilde{B}_{n}}) ||_2 \geq (\sigma_{re,n}+\sigma_{im,n})(\sqrt{a}+\sqrt{b}+\sqrt{2\ln(\frac{2N^2}{T})}) \}) \leq T \\
& \Rightarrow \mathbb{P}(\bigcap_{n} \{ || \text{Re}(\boldsymbol{\tilde{B}_{n}}) ||_2 + || \text{Im}(\boldsymbol{\tilde{B}_{n}}) ||_2 \leq (\sigma_{re,n}+\sigma_{im,n})(\sqrt{a}+\sqrt{b}+\sqrt{2\ln(\frac{2N^2}{T})}) \}) \geq 1-T \\
& \Rightarrow \mathbb{P}(|| \boldsymbol{U} ||_2 \leq \max_n [(\sigma_{re,n}+\sigma_{im,n})(\sqrt{a}+\sqrt{b}+\sqrt{2\ln(\frac{2N^2}{T})})] ) \geq 1-T \\
& \Rightarrow \mathbb{P}(|| \boldsymbol{U} ||_2 \leq 1.4q(\sqrt{a}+\sqrt{b}+\sqrt{2\ln(\frac{2N^2}{T})}) ) \geq 1-T \\
& \Leftrightarrow \mathbb{P}(||\boldsymbol{U}||_2 \geq 1.4q (\sqrt{a} + \sqrt{b}) + t    ) \leq 2N^2e^{-\frac{t^2}{2q^2}}
\end{split}
\end{equation}

\end{proof}

\subsection{Proof of Theorem 4.2}

\begin{proof}
We have to calculate the KL-term in Lemma 2.1 with the chosen distributions for $P$ and $\boldsymbol{u}$ ,for the value of 

\begin{equation}
\sigma  = \frac{\gamma}{42 B \tilde{\beta}^{d-1} [\sum_{i \in C} q_i \{\sqrt{a_i}+\sqrt{b_i}+\sqrt{2\log(4N_i^2d)} \} + \sum_{i \in D} \{2\sqrt{s_i}+\sqrt{2\log(2d)} \}]}. 
\end{equation}

We get:

\begin{equation}
KL(\boldsymbol{w}+\boldsymbol{u}||P) \leq \frac{|\boldsymbol{w}|^2}{2 \sigma^2} \leq \mathcal{O}(B^2 C_1^2 \frac { \prod_{i=1}^{k} ||\boldsymbol{W}_i||_2^2 }{ \gamma^2  } \sum_{i=1}^k \frac{||\boldsymbol{W}_i||_2^F}{||\boldsymbol{W}_i||_2^2} )
\end{equation}

with $C_1 = [\sum_{i \in C} q_i \{\sqrt{a_i}+\sqrt{b_i}+\sqrt{2\log(4N_i^2d)} \} + \sum_{i \in D} \{2\sqrt{s_i}+\sqrt{2\log(2d)} \}]$ where we have used the fact that both P and $\boldsymbol{u}$ follow multivariate Gaussian distributions. Theorem 4.2 results from substituting the value of $KL$ into Lemma 2.1. 

\begin{equation}
L_0(f_{\boldsymbol{w}}) \leq \hat{L}_{\gamma}(f_{\boldsymbol{w}})+\mathcal{O}(\sqrt{\frac{B^2 C_1^2 \prod_{i=1}^{k} ||\boldsymbol{W}_i||_2^2 \sum_{i=1}^k \frac{||\boldsymbol{W}_i||_2^F}{||\boldsymbol{W}_i||_2^2} +\text{ln} \frac{k m }{\delta} }{ \gamma^2 m  }}) 
\end{equation}

We now present a technical point regarding the parameter $\tilde{\beta}$. Recall that we mentioned in Lemma 2.2 that $\tilde{\beta}$ depends on a normalization of the network layers, we will formalise this concept below. The analysis is identical to the one in \citet{neyshabur2017pac}. 

Let $\beta = (\prod_{i=0}^d||\boldsymbol{W}_i||_2)^{1/d}$ and consider a network with the normalized weights $\tilde{\boldsymbol{W}}_i = \frac{\beta}{||\boldsymbol{W}_i||_2}\boldsymbol{W}_i$. Due to the homogeneity of the ReLu, we hvae that for feedforward neural networks with ReLu activations $f_{\tilde{\boldsymbol{w}}}=f_{\boldsymbol{w}}$ and so the (empirical and the expected) loss (including margin loss) is the same for $\tilde{\boldsymbol{w}} = \boldsymbol{w}$. We can also verify that $(\prod_{i=0}^d||\boldsymbol{W}_i||_2)=(\prod_{i=0}^d||\tilde{\boldsymbol{W}}_i||_2)$ and $\frac{||\boldsymbol{W}_i||_F}{||\boldsymbol{W}_i||_2}=\frac{||\tilde{\boldsymbol{W}}_i||_F}{||\tilde{\boldsymbol{W}}_i||_2}$, and so the excess error in the Theorem statement is also invariant to this transformation. It is therefore sufficient to prove the Theorem only for normalized weights $\tilde{\boldsymbol{w}}$, and hence we assume w.l.o.g. that the spectral norm is equal across layers, i.e. for any layer $i$, $||\boldsymbol{W}_i||_2$.

In the previous derivations we have set $\sigma$ according to $\beta$. More precisely, since the prior cannot depend on the learned predictor $\boldsymbol{w}$ or it's norm, we will set $\sigma$ based on an approximation $\tilde{\beta}$ For each value of $\tilde{\beta}$ on a pre-determined grid, we will compute the PAC-Bayes bound, establishing the generalization guaranteee for all $\boldsymbol{w}$ for which $|\beta-\tilde{\beta}|\leq\frac{1}{d}\beta$, and ensuring that each relevant value of $\beta$ is covered by some $\tilde{\beta}$ on the grid. We will then take a union bound over all $\tilde{\beta}$ on the grid. In the previous we have considered a fixed $\tilde{\beta}$ and the $\boldsymbol{w}$ for which $|\beta-\tilde{\beta}|\leq\frac{1}{d}\beta$, and hence $\frac{1}{e}\beta^{d-1} \leq \tilde{\beta}^{d-1} \leq e \beta^{d-1}$.

Finally we need to take a union bound over different choices of $\tilde{\beta}$. Let us see how many choices of $\tilde{\beta}$ we need to ensure we always have $\tilde{\beta}$ in the grid s.t. $|\beta-\tilde{\beta}|\leq\frac{1}{d}\beta$. We only need to consider values of $\beta$ in the range $(\frac{\gamma}{2B})^{1/d} \leq \beta \leq (\frac{\gamma\sqrt{m}}{2B})^{1/d}$. For $\beta$ outside this range the theorem statement holds trivially: Recall that the LHS of the theorem statement, $L_0(f_{\boldsymbol{w}})$ is always bounded by 1. If $\beta^d < \frac{\gamma}{2B}$, then for any $\boldsymbol{x}$, $|f_{\boldsymbol{w}}(\boldsymbol{x})| \leq \beta^dB \leq \gamma/2$ and therefore $L_{\gamma}=1$. Alternatively, if $\beta^d > \frac{\gamma\sqrt{m}}{2B}$, then the second term in equation (1) is greater than one. Hence, we only need to consider values of $\beta$ in the range discussed above. Since we need $\tilde{\beta}$ to satisfy $|\beta-\tilde{\beta}|\leq\frac{1}{d}\beta \leq \frac{1}{d}(\frac{\gamma}{2B})^{1/d}$, the size of the cover we need to consider is bounded by $dm^{\frac{1}{2d}}$. Taking a union bound over the choices of $\tilde{\beta}$ in this cover and using the bound in equation (33) gives us the theorem statement. 

\end{proof}

\end{document}